\documentclass[final,5p,times,twocolumn]{elsarticle}

\usepackage{comment}

\usepackage{wrapfig} 

\usepackage{graphicx,amssymb}

\usepackage{xcolor}

\usepackage{multirow}

\usepackage{subfig}

\usepackage[binary-units=true]{siunitx}

\usepackage{caption}

\usepackage{colortbl}

\definecolor{purple}{RGB}{128,0,128}

\definecolor{green}{RGB}{0,128,0}

\journal{Medical Image Analysis}

\begin{document}

\begin{frontmatter}

\title{Brain Tumor Segmentation with Deep Neural Networks\tnoteref{ti}}





\author[uds]{Mohammad Havaei\fnref{fn1}}

\author[ens]{Axel Davy} 

\author[udm]{David Warde-Farley} 

\author[udm,polyp]{Antoine Biard}   

\author[udm]{Aaron Courville}

\author[udm]{Yoshua Bengio}

\author[udm,poly]{Chris Pal}

\author[uds]{Pierre-Marc Jodoin}

\author[uds,twitter]{Hugo~Larochelle}


\address[uds]{Universit\'e de Sherbrooke, Sherbrooke, Qc, Canada}

\address[ens]{\'Ecole Normale sup\'erieure, Paris, France}

\address[udm]{Universit\'e de Montr\'eal, Montr\'eal, Canada}

\address[polyp]{\'Ecole polytechnique, Palaiseau, France}

\address[poly]{\'Ecole Polytechnique de Montr\'eal , Canada}
\address[twitter]{Twitter, USA}


\tnotetext[t1]{Accepted in Medical Image Analysis.}
\fntext[fn1]{mohammad.havaei@gmail.com}

\begin{abstract}

  In this paper, we present a fully automatic brain tumor segmentation method based on Deep Neural Networks (DNNs).  The proposed networks are tailored to glioblastomas (both low and high grade) pictured in MR images.  By their very nature, these tumors can appear anywhere in the brain and have almost any kind of shape, size, and contrast. These reasons motivate our exploration of a machine learning solution that exploits a flexible, high capacity DNN while being extremely efficient. Here, we give a description of different model choices that we've found to be necessary for obtaining competitive performance. We explore in particular different architectures based on Convolutional Neural Networks (CNN), i.e.\ DNNs specifically adapted to image data. 
  
  We present a novel CNN architecture which differs from those traditionally used in computer vision.  Our CNN exploits both local features as well as more global contextual features simultaneously.  Also, different from most traditional uses of CNNs, our networks use a final layer that is a convolutional implementation of a fully connected layer which allows a 40 fold speed up. We also describe a 2-phase training procedure that allows us to tackle difficulties related to the imbalance of tumor labels. Finally, we explore a cascade architecture in which the output of a basic CNN is treated as an additional source of information for a subsequent CNN. Results reported on the 2013 BRATS test dataset reveal that our architecture improves over the currently published state-of-the-art while being over 30 times faster.

\end{abstract}

\begin{keyword}


Brain tumor segmentation \sep deep neural networks


\end{keyword}

\end{frontmatter}


\section{Introduction}

In the United States alone, it is estimated that 23,000 new cases of brain cancer will be diagnosed in 2015\footnote{cancer.org}. While gliomas are the most common brain tumors, they can be less aggressive (i.e.\ low grade) in a patient with a life expectancy of several years, or more aggressive (i.e.\ high grade) in a patient with a life expectancy of at most 2 years.


Although surgery is the most common treatment for brain tumors, radiation and chemotherapy may be used to slow the growth of tumors that cannot be physically removed. Magnetic resonance imaging (MRI) provides detailed images of the brain, and is one of the most common tests used to diagnose brain tumors. All the more, brain tumor segmentation from MR images can have great impact for improved diagnostics, growth rate prediction and treatment planning.



While some tumors such as meningiomas can be easily segmented, others like gliomas and glioblastomas are much more difficult to localize.  These tumors (together with their surrounding edema) are often diffused, poorly contrasted, and extend tentacle-like structures that make them difficult to segment.  Another fundamental difficulty with segmenting brain tumors is that they can appear anywhere in the brain, in almost any shape and size.  
Furthermore, unlike images derived from X-ray computed tomography (CT) scans, the scale of voxel values in MR images is not standardized.
Depending on the type of MR machine used (1.5, 3 or 7 tesla) and the acquisition protocol (field of view value, voxel resolution, gradient strength, b0 value, etc.), the same tumorous cells may end up having drastically different grayscale values when pictured in different hospitals.

Healthy brains are typically made of 3 types of tissues: the white matter, the gray matter, and the cerebrospinal fluid.  The goal of brain tumor segmentation is to detect the location and extension of the tumor regions, namely active tumorous tissue (vascularized or not), necrotic tissue, and edema (swelling near the tumor). This is done by identifying abnormal areas when compared to normal tissue.  Since glioblastomas are infiltrative tumors, their borders are often fuzzy and hard to distinguish from healthy tissues.  As a solution, more than one MRI modality is often employed, e.g.\ T1 (spin-lattice relaxation), T1-contrasted (T1C), T2 (spin-spin relaxation), proton density (PD) contrast imaging, diffusion MRI (dMRI), and fluid attenuation inversion recovery (FLAIR) pulse sequences. The contrast between these modalities gives almost a unique signature to each tissue type.

 Most automatic brain tumor segmentation methods use hand-designed features~\citep{braintumorsegmentationdotorg,Menze2014}.  These methods implement a classical machine learning pipeline according to which features are first extracted and then given to a classifier whose training procedure does not affect the nature of those features. An alternative approach for designing task-adapted feature representations is to {\it learn} a hierarchy of increasingly complex features directly from in-domain data. Deep neural networks have been shown to excel at learning such feature hierarchies~\citep{bengio2013}.  In this work, we apply this approach to learn feature hierarchies adapted specifically to the task of brain tumor segmentation that combine information across MRI modalities.


Specifically, we investigate several choices for training Convolutional Neural Networks (CNNs), which are Deep Neural Networks (DNNs) adapted to image data. We report their advantages, disadvantages and performance using well established metrics. 
  Although CNNs first appeared over two decades ago~\citep{lecun1998},
they have recently become a mainstay of the computer vision community due to their record-shattering performance in the ImageNet Large-Scale Visual Recognition Challenge~\citep{Krizhevsky-2012-small}.
While CNNs have also been successfully applied to segmentation problems~\citep{Alvarez2012,long_shelhamer_fcn,SimulDetectSegm, ciresan2012}, most of the previous work has focused on non-medical tasks and many involve architectures that are not well suited to medical imagery or brain tumor segmentation in particular.  Our preliminary work on using convolutional neural networks for brain tumor segmentation together with two other methods using CNNs was presented in BRATS`14 workshop. However, those results were incomplete and required more investigation (More on this in chapter~\ref{sec:related_work}).


In this paper, we propose a number of specific CNN architectures for tackling brain tumor segmentation. Our architectures exploit the most recent advances in CNN design and training techniques, such as Maxout~\citep{Goodfellow_maxout_2013} hidden units and Dropout~\citep{Srivastava14a} regularization.  We also investigate several architectures which take into account both the local shape of tumors as well as their context.  

One problem with many machine learning methods is that they perform pixel classification without taking into account the local dependencies of labels (i.e.\ segmentation labels are conditionally independent given the input image). To account for this, one can employ structured output methods such as conditional random fields (CRFs), for which inference can be computationally expensive.  Alternatively, one can model label dependencies by considering the pixel-wise probability estimates of an initial CNN as additional input to certain layers of a second DNN, forming a cascaded architecture. Since convolutions are efficient operations, this approach can be significantly faster than implementing a CRF. 



We focus our experimental analysis on the fully-annotated MICCAI brain tumor segmentation (BRATS) challenge 2013 dataset~\citep{braintumorsegmentationdotorg} using the well defined training and testing splits, thereby allowing us to compare directly and quantitatively to a wide variety of other methods.

Our contributions in this work are four fold:
\begin{enumerate}
\item We propose a fully automatic method with results currently ranked second on the BRATS 2013 scoreboard; 
\item To segment a brain, our method takes between 25 seconds and 3 minutes, which is one order of magnitude faster than most state-of-the-art methods. 
\item Our CNN implements a novel two-pathway architecture that learns about the local details of the brain as well as the larger context.  We also propose a two-phase training procedure which we have found is critical to deal with imbalanced label distributions.  Details of these contributions are described in Sections~\ref{sec::twoPAths}~and~\ref{sec:training}.
\item We employ a novel cascaded architecture as an efficient and conceptually clean alternative to popular structured output methods.
Details on those models are presented in Section~\ref{sec::cascade}.
\end{enumerate}
%


\section{Related work}
\label{sec:related_work}

As noted by~\citet{Menze2014}, the number of publications devoted to automated brain tumor segmentation has grown exponentially in the last several decades. This observation 
not only underlines the need for automatic brain tumor segmentation tools, but also shows that research in that area is still a work in progress.

Brain tumor segmentation methods (especially those devoted to MRI) can be roughly divided in two categories: those based on generative models and those based on discriminative models~\citep{Menze2014,Bauer2013,Angelini2007}.

Generative models rely heavily on domain-specific prior knowledge about the appearance of both healthy and tumorous tissues. Tissue appearance is challenging to characterize, and existing generative models usually identify a tumor as being a shape or a signal which deviates from a normal (or average) brain~\citep{Clark1998}. Typically, these methods rely on anatomical models obtained after aligning the 3D MR image on an atlas or a template computed from several healthy brains~\citep{Doyle2013}. A typical generative model of MR brain images can be found in~\citet{Prastawa2004}. Given the ICBM brain atlas, the method aligns the brain to the atlas and computes posterior probabilities of healthy tissues (white matter, gray matter and cerebrospinal fluid) . Tumorous regions are then found by localizing voxels whose posterior probability is below a certain threshold. 
A post-processing step is then applied to ensure good spatial regularity. \citet{prastawa2003b} also register brain images onto an atlas in order to get a probability map for abnormalities.  An active contour 
is then initialized on this map and iterated until the change in posterior probability is below a certain threshold. Many other active-contour methods along the same lines have been proposed~\citep{Khotanlou2009,Cobzas2007,Popuri2012}, all of which depend on left-right brain symmetry features and/or alignment-based features.  
Note that since aligning a brain with a large tumor onto a template can be challenging, some methods perform registration and tumor segmentation at the same time~\citep{Kwon2014,Parisot2012}.

%

Other approaches for brain tumor segmentation employ discriminative models.  Unlike generative modeling approaches, these approaches exploit little prior knowledge on the brain's anatomy and instead rely mostly on the extraction of [a large number of] low level image features, directly modeling the relationship between these features and the label of a given voxel.  These features may be raw input pixels values~\citep{Havaei2014,hamamci2012a}, local histograms~\citep{Kleesiek2014,Meier2014} texture features such as Gabor filterbanks~\citep{Subbanna2013,Subbanna2014}, or alignment-based features such as inter-image gradient, region shape difference, and symmetry analysis~\citep{ANTsandArboles}.  Classical discriminative learning techniques such as SVMs~\citep{Bauer2011,schmidt2005,Lee2005} and decision forests~\citep{zikic2012} 
have also been used.  Results from the 2012, 2013 and 2014 editions of the MICCAI-BRATS Challenge suggest that methods relying on random forests are among the most accurate~\citep{Menze2014,Gotz2014,Kleesiek2014}.  

 One common aspect with discriminative models is their implementation of a conventional machine learning pipeline relying on hand-designed features. For these methods, the classifier is trained to separate healthy from non-heatlthy tissues assuming that the input features have a sufficiently high discriminative power since the behavior  the classifier is independent from nature of those features.  
One difficulty with methods based on hand-designed features is that they often require the computation of a large number of features in order to be accurate when used with many traditional machine learning techniques. This can make them slow to compute and expensive memory-wise.
More efficient techniques employ lower numbers of features, using dimensionality reduction or feature selection methods, but the reduction in the number of features is often at the cost of reduced accuracy.

By their nature, many hand-engineered features exploit very generic edge-related information, with no specific adaptation to the domain of brain tumors.  Ideally, one would like to have features that are composed and refined into higher-level, task-adapted representations. 
Recently, preliminary investigations have shown that the use of deep CNNs for brain tumor segmentation makes for a very promising approach (see the BRATS 2014 challenge workshop papers of \citet{Davy2014,Zikic2014,Urban2014}).   All three methods divide the 3D MR images into 2D~\citep{Davy2014,Zikic2014} or 3D patches~\citep{Urban2014} and train a CNN to predict its center pixel class. \citet{Urban2014} as well as \citet{Zikic2014} implemented a fairly common CNN, consisting of a series of convolutional layers, a non-linear activation function between each layer and a softmax output layer. Our work here\footnote{
It is important to note that while we did participate in the BRATS 2014 challenge, we could not report complete and fair experiments for it at the time of submitting this manuscript. See Section~\ref{sec:experiments} for a discussion on this point.
}
extends our preliminary results presented in \citet{Davy2014} using a two-pathway architecture, which we use here as a building block.  


In computer vision, CNN-based segmentation models have typically been applied to natural scene labeling.
For these tasks, the inputs to the model are the RGB channels of a patch from a color image. The work in \citet{pinheiro2014} uses a basic CNN to make predictions for each pixel and further improves the predictions by using them as extra information in the input of a second CNN model. 
Other work~\citep{farabet2013} involves several distinct CNNs processing the image at different resolutions. %
The final per-pixel class prediction is made by integrating information learned from all CNNs. To produce a smooth segmentation, these predictions are regularized using a more global superpixel segmentation of the image. Like our work, other recent work has exploited convolution operations in the final layer of a network to extend traditional CNN architectures for semantic scene segmentation \citep{long_shelhamer_fcn}. 
In the medical imaging domain in general there has been comparatively less work using CNNs for segmentation. However, some notable recent work by \citet{DeepAndWide2013} has used CNNs to predict the boundaries of neural tissue in electron microscopy images. Here we explore an approach with similarities to the various approaches discussed above, but 
in the context of brain tumor segmentation.

\section{Our Convolutional Neural Network Approach}
\label{sec:CNNApproach}

Since the brains in the BRATS dataset 
lack resolution in the third dimension, we consider performing the segmentation slice by slice from the axial view. 
Thus, our model processes sequentially each 2D axial image (slice) where each pixel is associated with different image modalities namely; T1, T2, T1C and FLAIR.  Like most CNN-based segmentation models~\citep{pinheiro2014,farabet2013}, our method predicts the class of a pixel by processing the $M\times M$ patch centered on that pixel.  The input ${\bf X}$ of our CNN model is thus an $M\times M$ 2D patch with several modalities.

%


The main building block used to construct a CNN architecture is the {\it convolutional layer}. Several layers can be stacked on top of each other forming a hierarchy of features. Each layer can be understood as extracting features from its preceding layer into the hierarchy to which it is connected. 
A single convolutional layer takes as input a stack of input planes and produces as output some number of output planes or \textit{feature maps}. Each feature map can be thought of as a topologically arranged map of responses of a particular spatially local non-linear feature extractor (the parameters of which are learned), applied identically to each spatial neighborhood of the input planes in a sliding window fashion.  In the case of a first convolutional layer, the individual input planes correspond to different MRI modalities (in typical computer vision applications, the individual input planes correspond to the red, green and blue color channels). In subsequent layers, the input planes typically consist of the feature maps of the previous layer.

Computing a feature map in a convolutional layer (see Figure~\ref{fig::convlayer} ) consists of the following three steps:

\begin{figure}

\centering

\includegraphics[width=0.75 \linewidth]{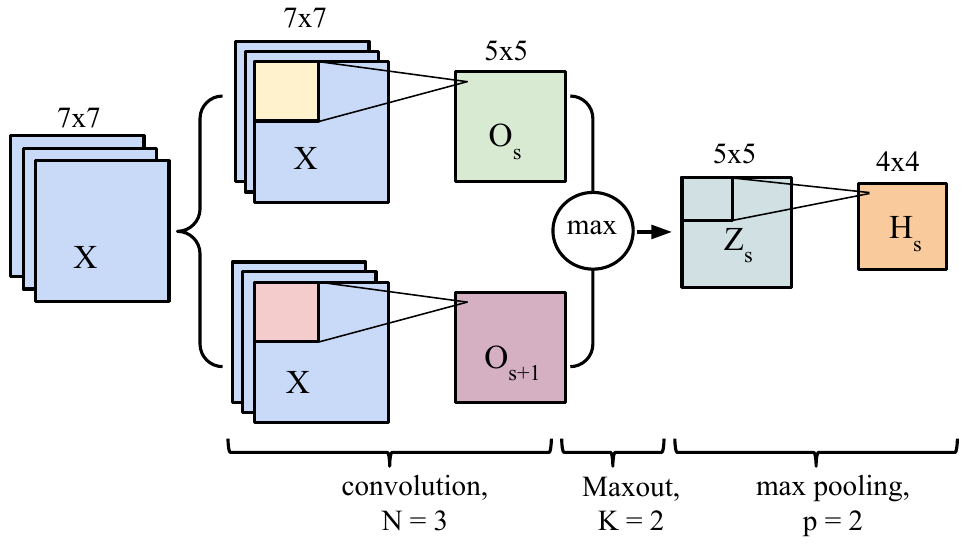}

\caption{A single convolution layer block showing computations for a single feature map. The input patch (here $7 \times 7$), is convolved with series of kernels (here $3 \times 3$) followed by Maxout and max-pooling. }

\label{fig::convlayer}
\end{figure}

\begin{enumerate}

\item {\it Convolution of kernels (filters):} Each feature map ${\bf O}_s$ is associated with one kernel (or several, in the case of Maxout). 
The feature map ${\bf O}_s$ is computed as follows:
%
\begin{equation}\label{eq:conv}
{\bf O}_s = b_s + \sum_r {\bf W}_{sr} \ast {\bf X}_r
\end{equation}


where ${\bf X}_r$ is the $r^{\rm th}$ input channel, ${\bf W}_{sr}$ is the sub-kernel for that channel, $\ast$ is the convolution operation and $b_s$ is a bias term\footnote{Since the convolutional layer is associated to $R$ input channels, ${\bf X}$ contains $M\times M\times R$ gray-scale values and thus each kernel  ${\bf W}_{s}$ contains $N\times N\times R$ weights. Accordingly, the number of parameters in a convolutional block of consisting of $S$ feature maps is equal to $R\times M\times M\times S $.}. In other words, the affine operation being performed for each feature map is the \textit{sum} of the application of $R$ different 2-dimensional $N\times N$ convolution filters (one per input channel/modality), plus a bias term which is added pixel-wise to each resulting spatial position. Though the input to this operation is a $M \times M \times R$ 3-dimensional tensor, the spatial topology being considered is 2-dimensional in the X-Y axial plane of the original brain volume.


Whereas traditional image feature extraction methods rely on a fixed recipe (sometimes taking the form of convolution with a linear e.g.\ Gabor filter bank), the key to the success of convolutional neural networks is their ability to learn the weights and biases of individual feature maps, giving rise to data-driven, customized, task-specific dense feature extractors. These parameters are adapted via stochastic gradient descent on a surrogate loss function related to the misclassification error, with gradients computed efficiently via the backpropagation algorithm~\citep{rumelhart1988learning}.

Special attention must be paid to the treatment of border pixels by the convolution operation. Throughout our architecture, we employ the so-called \textit{valid-mode} convolution, meaning that the filter response is not computed for pixel positions that are less than $\lfloor N/2\rfloor$ pixels away from the image border. An $N\times N$ filter convolved with an $M\times M$ input patch will result in a $Q \times Q$ output, where $Q = M - N + 1$. In Figure~\ref{fig::convlayer}, $M = 7$, $N = 3$ and thus $Q = 5$. 
Note that the size (spatial width and height) of the kernels are hyper-parameters that must be specified by the user. 

\item {\it Non-linear activation function:} To obtain features that are non-linear transformations of the input, an element-wise non-linearity is applied to the result of the kernel convolution. There are multiple choices for this non-linearity, such as the sigmoid, hyperbolic tangent and rectified linear functions~\citep{jarrett2009}, \citep{glorot2011}. 

Recently, \citet{Goodfellow_maxout_2013} proposed a Maxout non-linearity, which has been shown to be particularly effective at modeling useful features. Maxout features are associated with multiple kernels ${\bf W}_{s}$.  This implies each Maxout map ${\bf Z}_{s}$ is associated with $K$ feature maps : $\left \{ {\bf O}_s, {\bf O}_{s+1},...,{\bf O}_{s+K-1} \right \}$. Note that in Figure~\ref{fig::convlayer}, the Maxout maps are associated with $K = 2$ feature maps.  Maxout features correspond to taking the max over the feature maps ${\bf O}$, individually for each spatial position:

\begin{equation}
Z_{s,i,j} = \max \left \{ O_{s,i,j}, O_{s+1,i,j},..., O_{s+K-1,i,j} \right \}
\end{equation}
where $i,j$ are spatial positions. Maxout features are thus equivalent to using a convex activation function, but whose shape is adaptive and depends on the values taken by the kernels.

\item {\it Max pooling:} This operation consists of taking the maximum feature (neuron) value over sub-windows within each feature map. This can be formalized as follows:
\begin{equation}
H_{s,i,j} = \max_{p} Z_{s,i+p,j+p} , 
\end{equation} 
where $p$ determines the max pooling window size.  The sub-windows can be overlapping or not (Figure~\ref{fig::convlayer} shows an overlapping configuration). The max-pooling operation shrinks the size of the feature map. This is controlled by the pooling size $p$ and the stride hyper-parameter, which corresponds to the horizontal and vertical increments at which pooling sub-windows are positioned. Let $S$ be the stride value and $Q\times Q$ be the shape of the feature map before max-pooling. The output of the max-pooling operation would be of size $D \times D$, where $D = (Q-p)/S+1$.  In Figure~\ref{fig::convlayer}, since $Q=5, p=2, S=1$, the max-pooling operation results into a $D=4$ output feature map.
The motivation for this operation is to introduce invariance to local translations. This subsampling procedure has been found beneficial in other applications~\citep{Krizhevsky-2012-small}.

\end{enumerate}

Convolutional networks have the ability to extract a hierarchy of increasingly complex features which makes them very appealing. This is done by treating the output feature maps of a convolutional layer as input channels to the subsequent convolutional layer.

From the neural network perspective, feature maps correspond to a layer of hidden units or neurons. Specifically, each coordinate within a feature map corresponds to an individual neuron, for which the size of its receptive field corresponds to the kernel's size. A kernel's value also represents the weights of the connections between the layer's neurons and the neurons in the previous layer. It is often found in practice that the learned kernels resemble edge detectors, each kernel being tuned to a different spatial frequency, scale and orientation, as is appropriate for the statistics of the training data. 


Finally, to perform a prediction of the segmentation labels, we connect the last convolutional hidden layer to a convolutional output layer followed by a non-linearity (i.e.\ no pooling is performed).  It is necessary to note that, 
for segmentation purposes, a conventional CNN will not yield an efficient test time since the output layer is typically fully connected. By using a convolution at the end, for which we have an efficient implementation, the prediction at test time for a whole brain will be 45 times faster.
The convolution uses as many kernels as there are different segmentation labels (in our case five). Each kernel thus acts as the ultimate detector of tissue from one of the segmentation labels. We use the {\it softmax} non-linearity which normalizes the result of the kernel convolutions into a multinominal distribution over the labels. Specifically, let ${\bf a}$ be the vector of values at a given spatial position, it computes ${\rm softmax}({\bf a}) = \exp({\bf a})/Z$ where $Z=\sum_c \exp(a_c)$ is a normalization constant.  More details will be discussed in Section~\ref{sec:implementation}.

Noting ${\bf Y}$ as the segmentation label field over the input patch ${\bf X}$, we can thus interpret each spatial position of the convolutional output layer as providing a model for the likelihood distribution $p(Y_{ij}|{\bf X})$, where $Y_{ij}$ is the label at position $i,j$. We get the probability of all labels simply by taking the product of each conditional $p({\bf Y}|{\bf X}) = \prod_{ij} p(Y_{ij}|{\bf X})$. Our approach thus performs a multiclass labeling by assigning to each pixel the label with the largest probability.

\subsection{The Architectures}


Our description of CNNs so far suggests a simple architecture corresponding to a single stack of several convolutional layers. This configuration is the most commonly implemented architecture in the computer vision literature. However, one could imagine other architectures that might be more appropriate for the task at hand.

In this work, we explore a variety of architectures by using the concatenation of feature maps from different layers as another operation when composing CNNs. This operation allows us to construct architectures with multiple computational paths, which can each serve a different purpose.  We now describe the two types of architectures that we explore in this work.

\begin{figure*}
\centering
\includegraphics[width=0.75 \linewidth]{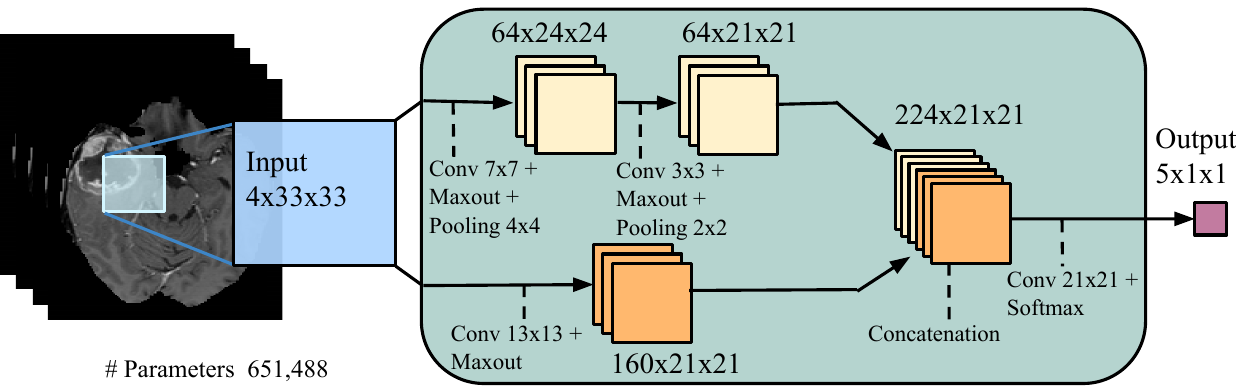}
\caption{Two-pathway CNN architecture (\textsc{TwoPathCNN}). The figure shows the input patch going through two paths of convolutional operations. The feature-maps in the local and global paths are shown in yellow and orange respectively. The convolutional layers used to produce these feature-maps are indicated by dashed lines in the figure. The green box embodies the whole model which in later architectures will be used to indicate the \textsc{TwoPathCNN}.}
\label{fig:basicmodel}
\end{figure*}

\subsubsection{Two-pathway architecture}
\label{sec::twoPAths}

This architecture is made of two streams: a pathway with smaller $7\times7$ receptive fields and another with larger $13\times13$ receptive fields. We refer to these streams as the {\em local} pathway and the {\em global} pathway, respectively. The motivation for this architectural choice is that we would like the prediction of the label of a pixel to be influenced by two aspects: the visual details of the region around that pixel and its larger ``context", i.e.\ roughly where the patch is in the brain. 

 The full architecture along with its details is illustrated in Figure~\ref{fig:basicmodel}. We refer to this architecture as the \textsc{TwoPathCNN}.
To allow for the concatenation of the top hidden layers of both pathways, we use two layers for the local pathway, with $3\times3$ kernels for the second layer. While this implies that the effective receptive field of features in the top layer of each pathway is the same, the global pathway's parametrization more directly and flexibly models features in that same area. The concatenation of the feature maps of both pathways is then fed to the output layer. 

\subsubsection{Cascaded architectures}
\label{sec::cascade}

One disadvantage of the CNNs described so far is that they predict each segmentation label separately from each other. This is unlike a large number of segmentation methods in the literature, which often propose a joint model of the segmentation labels, effectively modeling the direct dependencies between spatially close labels. One approach is to define a conditional random field (CRF) over the labels and perform mean-field message passing inference to produce a complete segmentation. In this case, the final label at a given position is effectively influenced by the models beliefs about what the label is in the vicinity of that position. 

On the other hand, inference in such joint segmentation methods is typically more computationally expensive than a simple feed-forward pass through a CNN. This is an important aspect that one should take into account if automatic brain tumor segmentation is to be used in a day-to-day practice.

Here, we describe CNN architectures that both exploit the efficiency of CNNs, while also more directly model the dependencies between adjacent labels in the segmentation. The idea is simple: since we'd like the ultimate prediction to be influenced by the model's beliefs about the value of nearby labels, we propose to feed the output probabilities of a first CNN as additional inputs to the layers of a second CNN. Again, we do this by relying on the concatenation of convolutional layers. In this case, we simply concatenate the output layer of the first CNN with any of the layers in the second CNN. Moreover, we use the same two-pathway structure for both CNNs. This effectively corresponds to a cascade of two CNNs, thus we refer to such models as cascaded architectures.

\begin{figure*}[htp]

\centering

\subfloat[Cascaded architecture, using input concatenation (\textsc{InputCascadeCNN}).]{\label{fig:traf-a}\includegraphics[width=0.95 \linewidth]{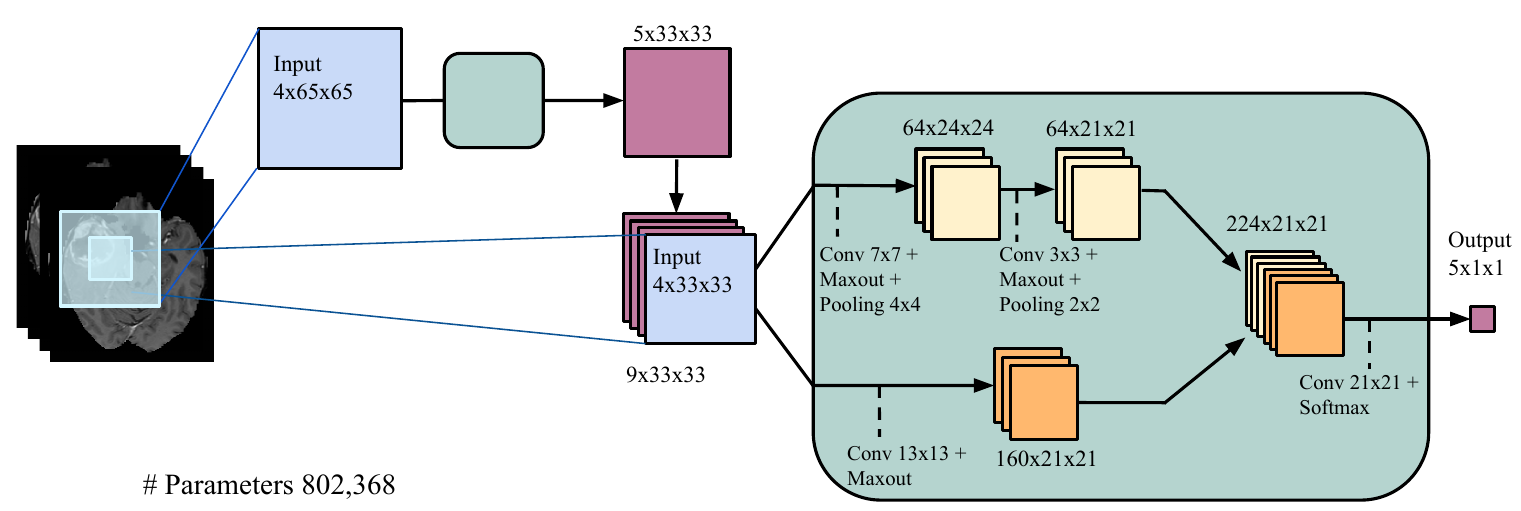}}
\newline
\subfloat[Cascaded architecture, using local pathway concatenation (\textsc{LocalCascadeCNN}).]{\label{fig:traf-b}\includegraphics[width=0.95 \linewidth]{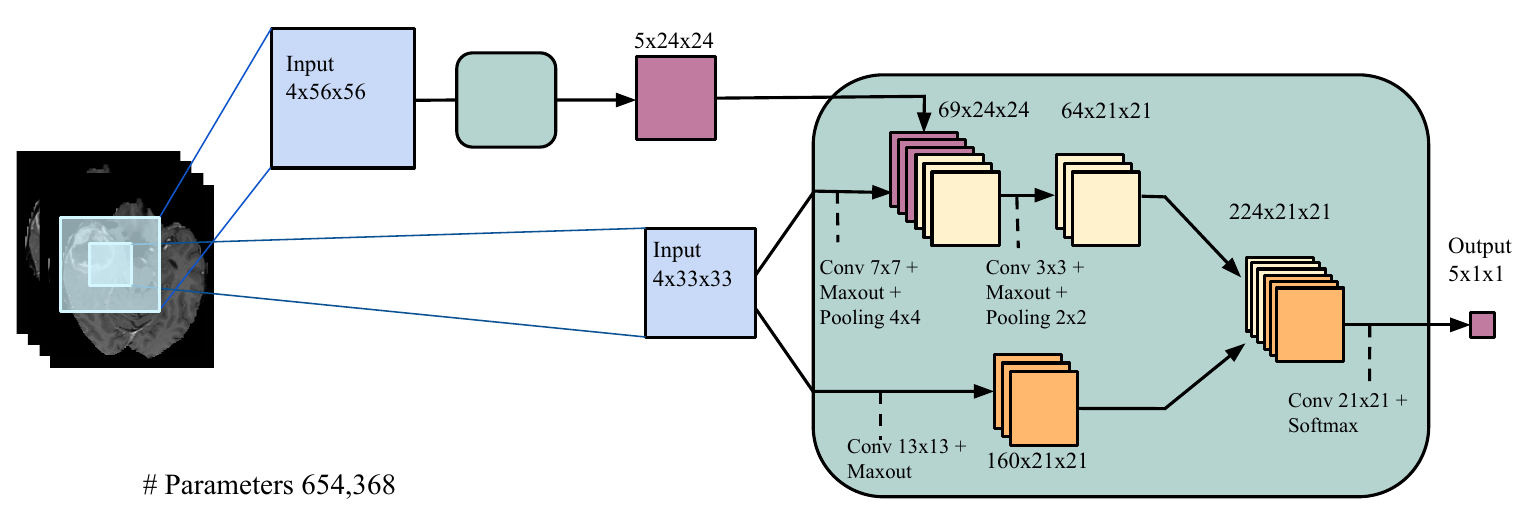}}
\newline
\subfloat[Cascaded architecture, using pre-output concatenation, which is an architecture with properties similar to that of learning using a limited number of mean-field inference iterations in a CRF (\textsc{MFCascadeCNN}).]{\label{fig:traf-c}\includegraphics[width=0.95 \linewidth]{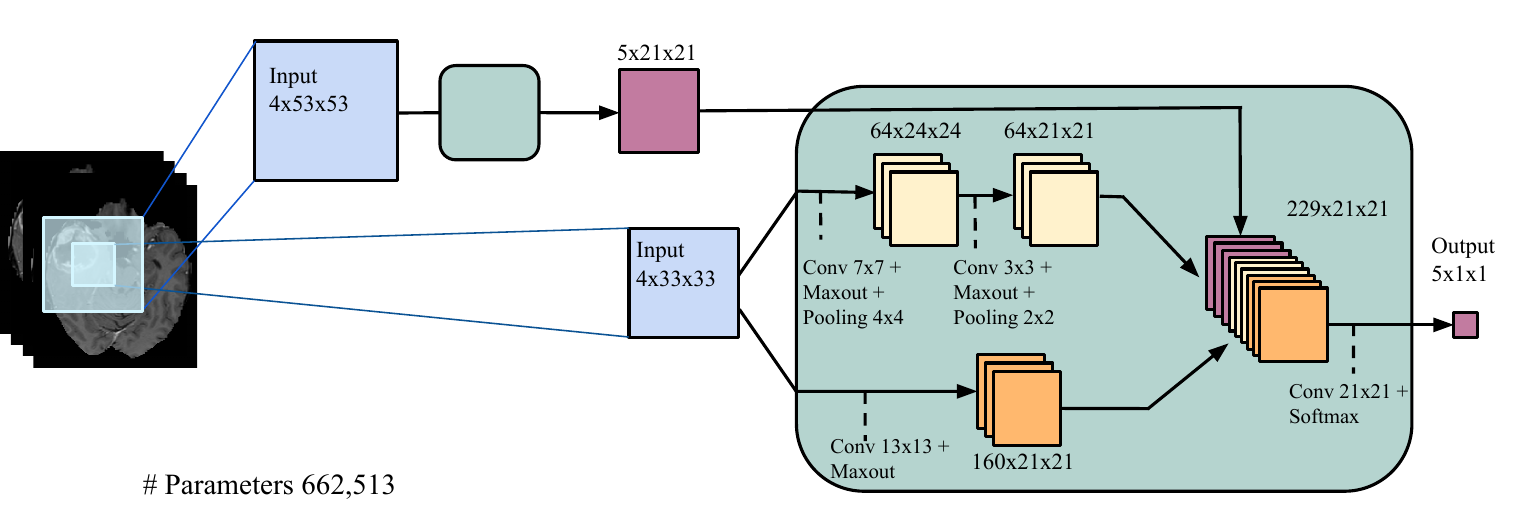}}

\caption{Cascaded architectures.}

\label{fig:traf}

\end{figure*}

In this work, we investigated three cascaded architectures that concatenate the first CNN's output at different levels of the second CNN:

\begin{itemize}

\item {\it Input concatenation:} In this architecture, we provide the first CNN's output directly as input to the second CNN. They are thus simply treated as additional image channels of the input patch. The details are illustrated in Figure~\ref{fig:traf-a}. We refer to this model as \textsc{InputCascadeCNN}.

\item {\it Local pathway concatenation:} In this architecture, we move up one layer in the local pathway and perform concatenation to its first hidden layer, in the second CNN. The details are illustrated in Figure~\ref{fig:traf-b}. We refer to this model as  \textsc{LocalCascadeCNN}. 

\item {\it Pre-output concatenation:} In this last architecture, we move to the very end of the second CNN and perform concatenation right before its output layer. This architecture is interesting, as it is similar to the computations made by one pass of mean-field inference~\citep{xing2002} in a CRF whose pairwise potential functions are the weights in the output kernels. From this view, the output of the first CNN is the first iteration of mean-field, while the output of the second CNN would be the second iteration. The difference with regular mean-field however is that our CNN allows the output at one position to be influenced by its previous value, and the convolutional kernels are not the same in the first and second CNN. The details are illustrated in Figure~\ref{fig:traf-c}. We refer to this model as \textsc{MFCascadeCNN}.

\end{itemize}

\subsection{Training}
\label{sec:training}

\paragraph{Gradient Descent}

By interpreting the output of the convolutional network as a model for the distribution over segmentation labels, a natural training criteria is to maximize the probability of all labels in our training set or, equivalently, to minimize the negative log-probability $-\log p({\bf Y}|{\bf X}) = \sum_{ij} -\log p(Y_{ij}|{\bf X})$ for each labeled brain.

To do this, we follow a stochastic gradient descent approach by repeatedly selecting labels $Y_{ij}$ at a random subset of patches within each brain, computing the average negative log-probabilities for this mini-batch of patches and performing a gradient descent step on the CNNs parameters (i.e.\ the kernels at all layers).

Performing updates based only on a small subset of patches allows us to avoid having to process a whole brain for each update, while providing reliable enough updates for learning. In practice, we implement this approach by creating a dataset of mini-batches of smaller brain image patches, paired with the corresponding center segmentation label as the target.

To further improve optimization, we implemented a so-called {\em momentum} strategy which has been shown successful in the past~\citep{Krizhevsky-2012-small}.  The idea of momentum is to use a temporally averaged gradient in order to damp the optimization velocity:

\begin{eqnarray}
{\bf V}_{i+1} & = & \mu * {\bf V}_i - \alpha * \nabla {\bf W}_i \nonumber \\
{\bf W}_{i+1} & = & {\bf W}_{i}+{\bf V}_{i+1} \nonumber
\end{eqnarray}

where ${\bf W}_i$ stands for the CNNs parameters at iteration $i$, $ \nabla{\bf W}_i$ the gradient of the loss function at ${\bf W}_i$, {\bf V} is the integrated velocity initialized at zero, $\alpha$ is the learning rate, and $\mu$ the momentum coefficient.  We define a schedule for the  momentum $\mu$ where the momentum coefficient is gradually increased during training. In our experiments the initial momentum coefficient was set to $\mu=0.5$ and the final value was set to $\mu=0.9$.

Also, the learning rate $\alpha$ is decreased by a factor at every epoch. The initial learning rate was set to $\alpha=0.005$ and the decay factor to $10^{-1}$. 


\paragraph{Two-phase training}

\label{twopahsetraining}

Brain tumor segmentation is a highly data imbalanced problem where the healthy voxels (i.e.\ label 0) comprise 98\% of total voxels. From the remaining 2\% pathological voxels,  0.18\% belongs to necrosis (label 1), 1.1\% to edema (label 2), 0.12\% to non-enhanced (label 3) and 0.38\% to enhanced tumor (label 4). Selecting patches from the true distribution would cause the model to be overwhelmed by healthy patches and causing problem when training out CNN models. Instead, we initially construct our patches dataset such that all labels are equiprobable. This is what we call the {\em first} training phase.  Then, in a {\em second} phase, we  account for the un-balanced nature of the data and re-train only the output layer (i.e.\ keeping the kernels of all other layers fixed) with a more representative distribution of the labels. This way we get the best of both worlds: most of the capacity (the lower layers) is used in a balanced way to account for the diversity in all of the classes, while the output probabilities are calibrated correctly (thanks to the re-training of the output layer with the natural frequencies of classes in the data).

\paragraph{Regularization}

Successful CNNs tend to be models with a lot of capacity, making them vulnerable to overfitting in a setting like ours where there clearly are not enough training examples.   Accordingly, we found that regularization is important in obtaining good results. Here, regularization took several forms. First, in all layers, we bounded the absolute value of the kernel weights and applied both L1 and L2 regularization to prevent overfitting. This is done by adding the regularization terms to the negative log-probability (i.e.\ $-\log p({\bf Y}|{\bf X}) + \lambda_1 \|{\bf W}\|_1 + \lambda_2\|{\bf W}\|^2$, where $\lambda_1$ and $\lambda_2$ are coefficients for L1 and L2 regularization terms respectively). L1 and L2 affect the parameters of the model in different ways, while L1 encourages sparsity, L2 encourages small values. 
We also used a validation set for early stopping, i.e.\ stop training when the validation performance stopped improving. The validation set was also used to tune the other hyper-parameters of the model. The reader shall note that the hyper-parameters of the model which includes using or not L2 and/or L1 coefficients were selected by doing a grid search over range of parameters.  The chosen hyper-parameters were the ones for which the model performed best on a validation set.

Moreover, we used {\it Dropout}~\citep{Srivastava14a}, a recent regularization method that works by stochastically adding noise in the computation of the hidden layers of the CNN. This is done by multiplying each hidden or input unit by 0 (i.e.\ masking) with a certain probability (e.g.\ 0.5), independently for each unit and training update. This encourages the neural network to learn features that are useful ``on their own", since each unit cannot assume that other units in the same layer won't be masked as well and co-adapt its behavior. At test time, units are instead multiplied by one minus the probability of being masked. For more details, see~\citet{Srivastava14a}.

 Considering the large number of parameters our model has, one might think that even with our regularization strategy, the 30 training brains from BRATS 2013 are too few to prevent overfitting.  But as will be shown in the results section, our model generalizes well and thus do not overfit.  One reason for this is the fact that each brain comes with 200 2d slices and thus, our model has approximately 6000 2D images to train on.   We shall also mention that by their very nature, MRI images of brains are very similar from one patient to another. Since the variety of those images is much lower than those in real-image datasets such as CIFAR and ImageNet, a fewer number of training samples is thus needed.

\paragraph{Cascaded Architectures}

To train a cascaded architecture, we start by training the \textsc{TwoPathCNN} with the two phase stochastic gradient descent procedure described previously. Then, we fix the parameters of the \textsc{TwoPathCNN} and include it in the cascaded architecture (be it the \textsc{InputCascadeCNN}, the \textsc{LocalCascadeCNN}, or the \textsc{MFCascadeCNN}) and move to training the remaining parameters using a similar procedure. It should be noticed however that for the spatial size of the first CNN's output and the layer of the second CNN to match, we must feed to the first CNN a much larger input. Thus, training of the second CNN must be performed on larger patches. For example in the \textsc{InputCascadeCNN} (Figure \ref{fig:traf-a}), the input size to the first model is of size $65\times65$ which results into an output of size $33\times33$. Only in this case the outputs of the first CNN can be concatenated with the input channels of the second CNN. 



\section{Implementation details}

\label{sec:implementation}

Our implementation is based on the Pylearn2 library~\citep{pylearn2_arxiv_2013}. Pylearn2 is an open-source machine learning library specializing in deep learning algorithms. It also supports the use of GPUs, which can greatly accelerate the execution of deep learning algorithms.

Since CNN's are able to learn useful features from scratch, we applied only minimal pre-processing. We  employed the same pre-processing as Tustison et al., the winner of the 2013 BRATS challenge~\citep{Menze2014}. The pre-processing follows three steps. First, the 1\% highest and lowest intensities are removed. Then, we apply an N4ITK bias correction~\citep{avants2009advanced} to T1 and T1C modalities. 
The data is then normalized within each input channel by subtracting the channel's mean and dividing by the channel's standard deviation.

As for post-processing, a simple method based on connected components was implemented to remove flat blobs which might appear in the predictions due to bright corners of the brains close to the skull.

The hyper-parameters of the different architectures (kernel and max pooling size for each layer and the number of layers) can be seen in Figure~\ref{fig:traf}. Hyper-parameters were tuned using grid search and cross-validation on a validation set (see~\citet{bengio2012}). The chosen hyper-parameters were the ones for which the model performed best on the validation set. For max pooling, we always use a stride of 1. This is to keep per-pixel accuracy during full image prediction. We observed in practice that max pooling in the global path does not improve accuracy.  We also found that adding additional layers to the architectures or increasing the capacity of the model by adding additional feature maps to the convolutional blocks do not provide any meaningful performance improvement. 

Biases are initialized to zero except for the softmax layer for which we initialized them to the $\log$ of the label frequencies. The kernels are randomly initialized from $U\left(-0.005 , 0.005\right)$.
Training takes about 3 minutes per epoch for the \textsc{TwoPathCNN} model on an NVIDIA Titan black card.

At test time, we run our code on a GPU in order to exploit its computational speed. Moreover, the convolutional nature of the output layer allows us to further accelerate computations at test time. This is done by feeding as input a full image and not individual patches.  Therefore, convolutions at all layers can be extended to obtain all label probabilities $p(Y_{ij}|{\bf X})$ for the entire image.
With this implementation, we are able to produce a segmentation in 25 seconds per brain on the Titan black card with the \textsc{TwoPathCNN} model. This turns out to be 45 times faster than when we extracted a patch at each pixel and processed them individually for the entire brain. 

Predictions for the \textsc{MFCascadeCNN} model, the \textsc{LocalCascadeCNN} model, and  \textsc{InputCascadeCNN} model take on average 1.5 minutes, 1.7 minutes and 3 minutes respectively.

\section{Experiments and Results}
\label{sec:experiments}

\begin{figure}
\centering
\includegraphics[width=\linewidth]{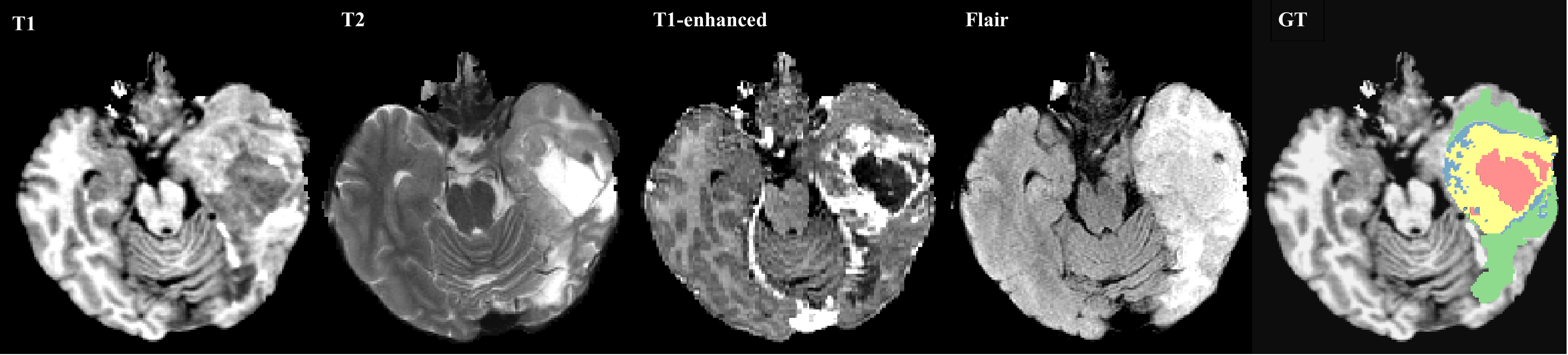}
\caption{The first four images from left to right show the MRI modalities used as input channels to various CNN models and the fifth image shows the ground truth labels where \textcolor[RGB]{135,213,120}{$\blacksquare$} edema, \textcolor[RGB]{225,225,95}{$\blacksquare$} enhanced tumor,
\textcolor[RGB]{246,145,139}{$\blacksquare$} necrosis,
\textcolor[RGB]{124,167,208}{$\blacksquare$} non-enhanced tumor.}
\label{fig:data}
\end{figure}

The experiments were carried out on real patient data obtained from the 2013 brain tumor segmentation challenge (BRATS2013), as part of the MICCAI conference~\citep{braintumorsegmentationdotorg}. The BRATS2013 dataset is comprised of 3 sub-datasets. The training dataset, which contains 30 patient subjects all with pixel-accurate ground truth (20 high grade and 10 low grade tumors); the test dataset which contains 10 (all high grade tumors) and the leaderboard dataset which contains 25 patient subjects (21 high grade and 4 low grade tumors). There is no ground truth provided for the test and leaderboard datasets. All brains in the dataset have the same orientation. For each brain there exists 4 modalities, namely T1, T1C, T2 and Flair which are co-registered. The training brains come with groundtruth for which 5 segmentation labels are provided, namely {\it non-tumor}, {\it  necrosis}, {\it edema}, {\it non-enhancing tumor} and {\it enhancing tumor}. Figure~\ref{fig:data} shows an example of the data as well as the ground truth. In total, the model iterates over about 2.2 million examples of tumorous patches (this consists of all the 4 sub-tumor classes) and goes through 3.2 million of the healthy patches. As mentioned before during the first phase training, the distribution of examples introduced to the model from all 5 classes is uniform.

Please note that we could not use the BRATS 2014 dataset due to problems with both the system performing the evaluation and the quality of the labeled data. For these reasons the old BRATS 2014 dataset has been removed from the official website and, at the time of submitting this manuscript, the BRATS website still showed: ``Final data for BRATS 2014 to be released soon''. Furthermore, we have even conducted an experiment where we trained our model with the old 2014 dataset and made predictions on the 2013 test dataset; however, the performance was worse than our results mentioned in this paper. For these reasons, we decided to focus on the BRATS 2013 data.

As mentioned in Section~\ref{sec:CNNApproach}, we work with 2D slices due to the fact that the MRI volumes in the dataset do not posses an isotropic resolution and the spacing in the third dimension is not consistent across the data. We explored the use of 3D information (by treating the third dimension as extra input channels or by having an architecture which takes orthogonal slices from each view and makes the prediction on the intersecting center pixel), but that didn't improve performance and made our method very slow.

Note that as suggested by \citet{Krizhevsky-2012-small}, we applied data augmentation by flipping the input images.  Unlike what was reported by ~\citet{zeiler2014},  it did not improve the overall accuracy of our model.

 Quantitative evaluation of the models performance on the test set is achieved by uploading the segmentation results to the online BRATS evaluation system~\citep{BRATSURL}. The online system provides the quantitative results as follows: 
The tumor structures are grouped in 3 different tumor regions. This is mainly due to practical clinical applications. 
As described by~\citet{Menze2014}, tumor regions are defined as:

\begin{enumerate}[a)]

{\setlength\itemindent{25pt} \item The {\it complete} tumor region (including all four tumor structures).}

{\setlength\itemindent{25pt}\item The {\it core} tumor region (including all tumor structures exept ``edema").}

{\setlength\itemindent{25pt}\item The {\it enhancing} tumor region (including the ``enhanced tumor" structure).}

\end{enumerate} 

For each tumor region, {\it Dice} (identical to F measure), {\it Sensitivity} and {\it Specificity} are computed as follows :

\begin{eqnarray}
Dice(P,T) &=& \frac{|P_1 \wedge T_1|}{(|P_1|+|T_1|)/2}, \nonumber \\
Sensitivity(P,T) &=& \frac{|P_1 \wedge T_1|}{|T_1|}, \nonumber \\
Specificity(P,T) &=& \frac{|P_0 \wedge T_0|}{|T_0|}, \nonumber
\end{eqnarray}

where $P$ represents the model predictions and $T$ represents the ground truth labels. We also note as $T_1$ and $T_0$ the subset of voxels predicted as positives and negatives for the tumor region in question. Similarly for $P_1$ and $P_0$.  The online evaluation system also provides a ranking for every method submitted for evaluation. This includes methods from the 2013 BRATS challenge published in \citep{Menze2014} as well as anonymized unpublished methods for which no reference is available. In this section, we report experimental results for our different CNN architectures.

\subsection{The \textsc{TwoPathCNN} architecture}

As mentioned previously, unlike conventional CNNs, the \textsc{TwoPathCNN} architecture has two pathways: a ``local" path focusing on details and a ``global" path more focused on the context. To better understand how joint training of the global and local pathways benefits the performance, we report results on each pathway as well as results on averaging the outputs of each pathway when trained separately. Our method also deals with the unbalanced nature of the problem by training in two phases as discussed in Section \ref{twopahsetraining}. To see the impact of the two phase training, we report results with and without it. We refer to the CNN model consisting of only the local path (i.e.\ conventional CNN architecture) as \textsc{LocalPathCNN}, the CNN model consisting of only the global path as \textsc{GlobalPathCNN}, the model averaging the outputs of the local and global paths (i.e.\ \textsc{LocalPathCNN} and \textsc{GlobalPathCNN}) as \textsc{AverageCNN} and the two-pathway CNN architecture as \textsc{TwoPathCNN}. The second training phase is noted by appending `*' to the architecture name. Since the second phase training has a substantial effect and always improves the performance, we only report results on  \textsc{GlobalPathCNN} and \textsc{AverageCNN} with the second phase.  

Table~\ref{tab:basicmodel} presents the quantitative results of these variations. This table contains results for the \textsc{TwoPathCNN} with one and two training phases, the common single path CNN (i.e.\  \textsc{LocalPathCNN}) with one and two training phases, the \textsc{GlobalPathCNN*} which is a single path CNN model following the global pathway architecture and the output average of each of the trained single-pathway models (\textsc{AverageCNN*}). Without much surprise, the single path with one training phase CNN was ranked last with the lowest scores on almost every region.  Using a second training phase gave a significant boost to that model with a rank that went from 15 to 9. Also, the table shows that joint training of the local and global paths yields better performance compared to when each pathway is trained separately and the outputs are averaged. One likely explanation is that by joint training the local and global paths, the model allows the two pathways to co-adapt. In fact, the \textsc{AverageCNN*} performs worse than the \textsc{LocalPathCNN*} due to the fact that the \textsc{GlobalPathCNN*} performs very badly. The top performing method in the uncascaded models is the \textsc{TwoPathCNN*} with a rank of 4.

Also, in some cases results are less accurate over the Enhancing region than for the Core and Complete regions.  There are 2 main reasons for that.  First, borders are usually diffused and there are no clear cut between enhanced tumor and non-enhanced tissues. This creates problems for both user labeling, ground truth, as well as the model. The second reason is that the model learns what it sees in the ground truth. Since the labels are created by different people and since the borders are not clear, each user has a slightly different interpretation of the borders of the enhanced tumor and so sometimes we see overly thick enhanced tumor in the ground truth.

\begin{figure}

\centering

\includegraphics[width=\linewidth]{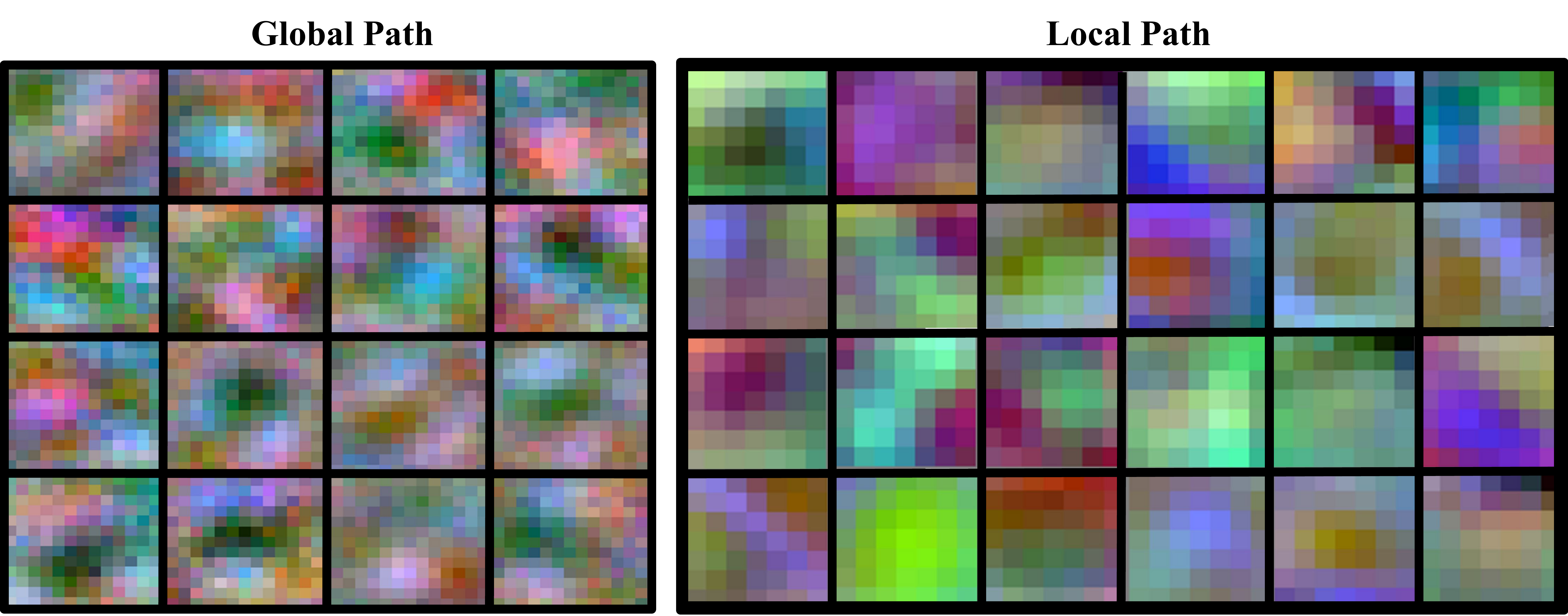}

\caption{Randomly selected filters from the first layer of the model. From left to right the figure shows visualization of features from the first layer of the global and local path respectively. Features in the local path include more edge detectors while the global path contains more localized features.}

\label{fig:features}
\end{figure}

\begin{figure*}[]
\centering

\includegraphics[width=\linewidth]{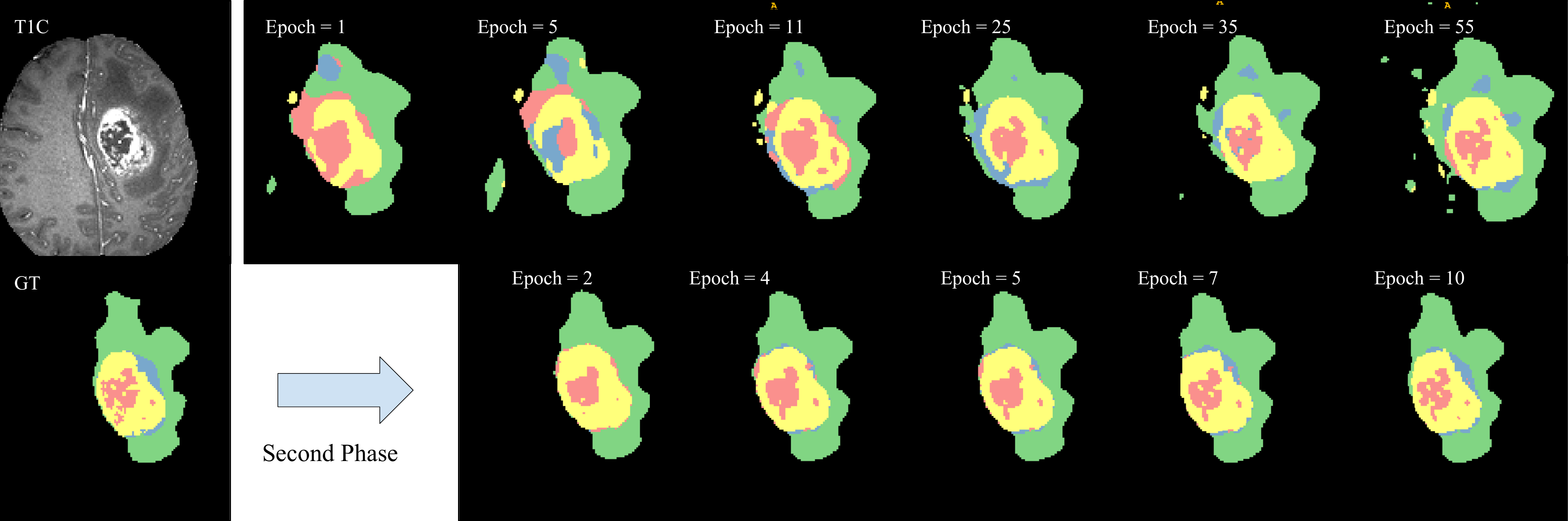}

\caption{Progression of learning in \textsc{InputCascadeCNN*}. The stream of figures on the first row from left to right show the learning process during the first phase. As the model learns better features, it can better distinguish boundaries between tumor sub-classes. This is made possible due to uniform label distribution of patches during the first phase training which makes the model believe all classes are equiprobable and causes some false positives. This drawback is alleviated by training a second phase (shown in second row from left to right) on a distribution closer to the true distribution of labels. The color codes are as follows:  \textcolor[RGB]{135,213,120}{$\blacksquare$} edema, \textcolor[RGB]{225,225,95}{$\blacksquare$} enhanced tumor,
\textcolor[RGB]{246,145,139}{$\blacksquare$} necrosis,
\textcolor[RGB]{124,167,208}{$\blacksquare$} non-enhanced tumor.}
\label{fig:scheduled_prediction}
\end{figure*}

{Figure~\ref{fig:features} shows representation of low level features in both local and global paths. As seen from this figure,  features in the local path include more edge detectors while the ones in the global path are more localized features. Unfortunately, visualizing the learned mid/high level features of a CNN is still very much an open research problem. However, we can study the impact these features have on predictions by visualizing the segmentation results of different models.
The segmentation results on two subjects from our validation set, produced by different variations of the basic model can be viewed in Figure~\ref{fig:results}\footnote{It is important to note that we do not train the model on the validation set and thus the quality of the results is not due to overfitting}. As shown in the figure, the two-phase training procedure allows the model to learn from a more realistic distribution of labels and thus removes false positives produced by the model which trains with one training phase.  Moreover, by having two pathways, the model can simultaneously learn the global contextual features as well as the local detailed features. This gives the advantage of correcting labels at a global scale as well as recognizing fine details of the tumor at a local scale, yielding a better segmentation as oppose to a single path architecture which results in smoother boundaries. 
Joint training of the two convolutional pathways and having two training phases achieves better results.

\begin{figure*}[p!]

\centering

\includegraphics[width=\linewidth]{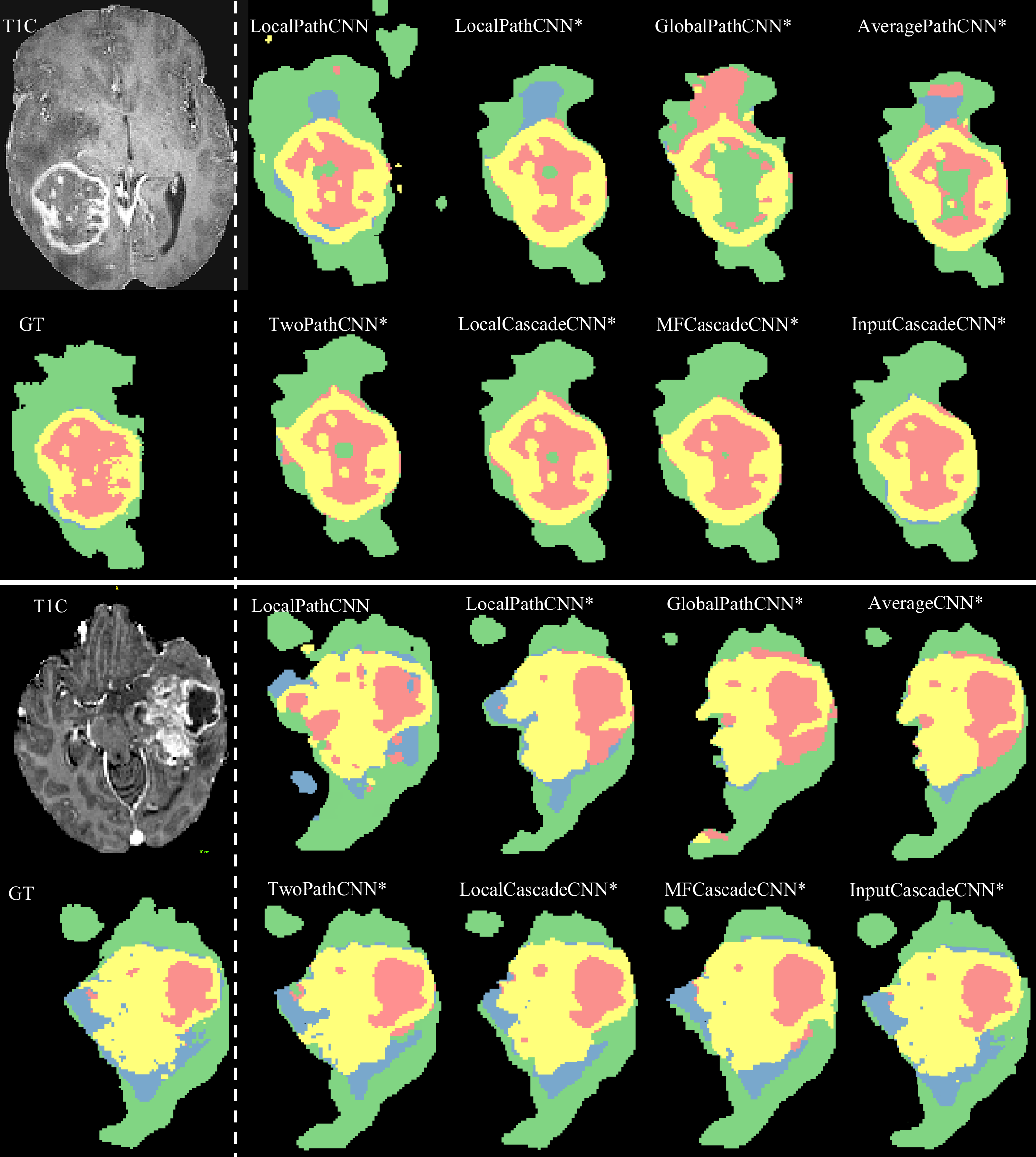}

\caption{Visual results from our CNN architectures from the Axial view. For each sub-figure, the top row from left to right shows T1C modality, the conventional one path CNN, the Conventional CNN with two training phases, and the \textsc{TwoPathCNN} model.  The second row from left to right shows the ground truth,  \textsc{LocalCascadeCNN} model, the \textsc{MFCascadeCNN} model and the \textsc{InputCascadeCNN}. The color codes are as follows:  \textcolor[RGB]{135,213,120}{$\blacksquare$} edema, \textcolor[RGB]{225,225,95}{$\blacksquare$} enhanced tumor,
\textcolor[RGB]{246,145,139}{$\blacksquare$} necrosis,
\textcolor[RGB]{124,167,208}{$\blacksquare$} non-enhanced tumor.}
\label{fig:results}
\end{figure*}

\begin{figure}[tp]

\centering

\includegraphics[width=0.8\linewidth]{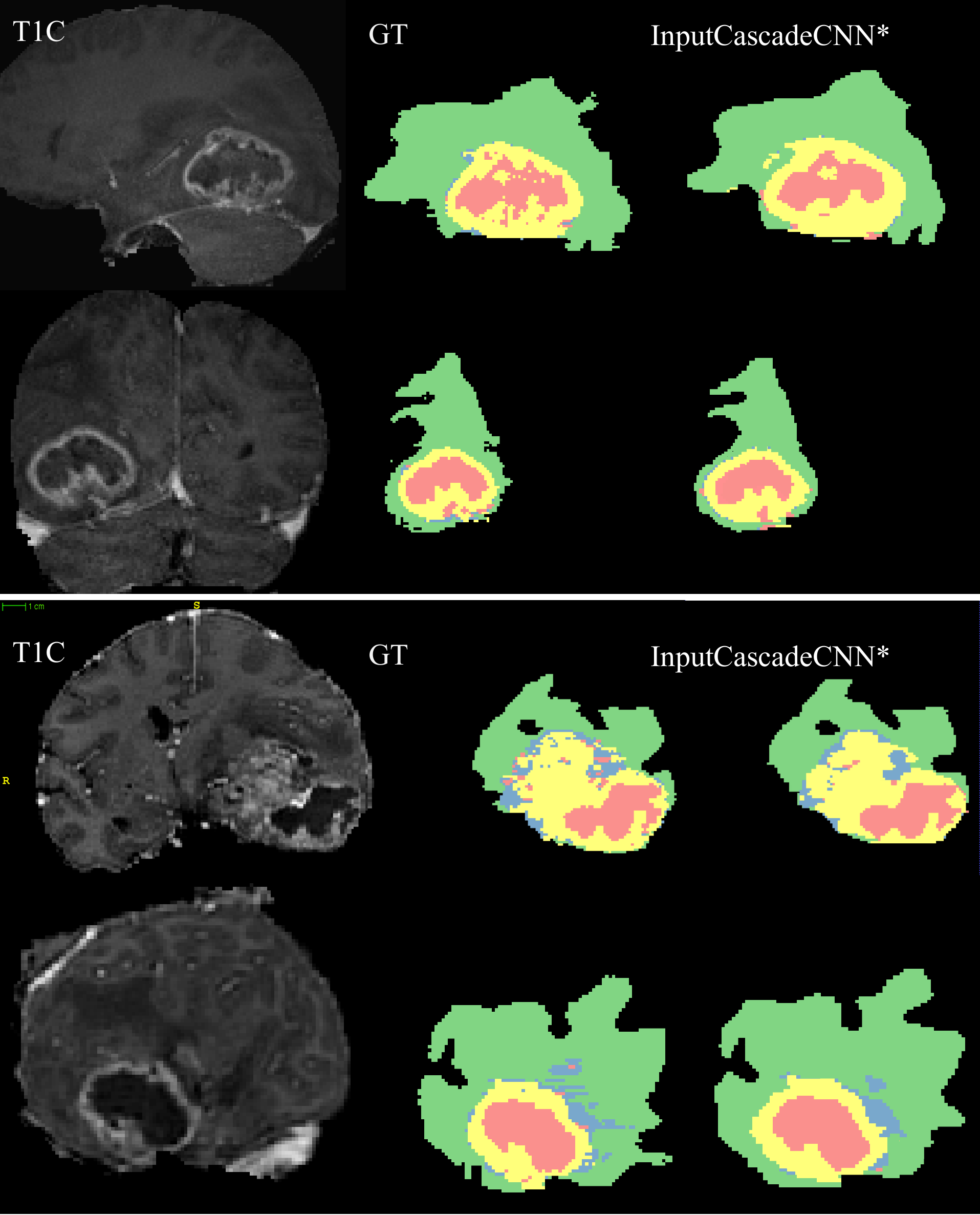}

\caption{Visual results from our top performing model, \textsc{InputCascadeCNN*} on Coronal and Sagittal views. The subjects are the same as in Figure~\ref{fig:results}. In every sub-figure, the top row represents the Sagital view and the bottom row represents the Coronal view. The color codes are as follows:  \textcolor[RGB]{135,213,120}{$\blacksquare$} edema, \textcolor[RGB]{225,225,95}{$\blacksquare$} enhanced tumor,
\textcolor[RGB]{246,145,139}{$\blacksquare$} necrosis,
\textcolor[RGB]{124,167,208}{$\blacksquare$} non-enhanced tumor.}
\label{fig:multiple_views}
\end{figure}

\begin{figure}[tp]
\centering

\includegraphics[width=0.8\linewidth]{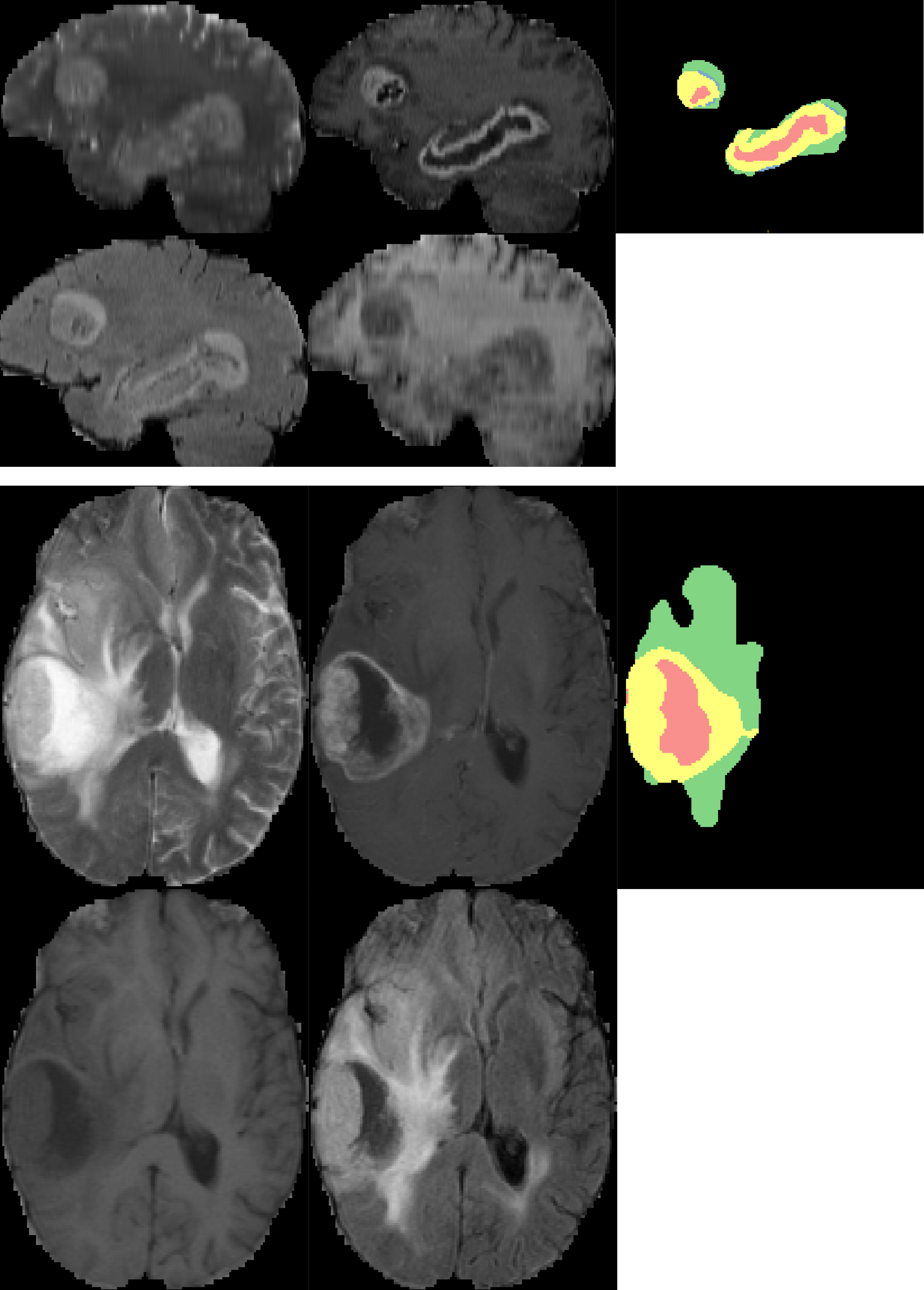}

\caption{Visual segmentation results from our top performing model, \textsc{InputCascadeCNN*}, on examples of the BRATS2013 test dataset in Saggital (top) and Axial (bottom) views. The color codes are as follows:  \textcolor[RGB]{135,213,120}{$\blacksquare$} edema, \textcolor[RGB]{225,225,95}{$\blacksquare$} enhanced tumor,
\textcolor[RGB]{246,145,139}{$\blacksquare$} necrosis,
\textcolor[RGB]{124,167,208}{$\blacksquare$} non-enhanced tumor.}
\label{fig:testresults}
\end{figure} 

\begin{table*}[tp]
\begin{center}
\caption{Performance of the \textsc{TwoPathCNN} model and variations. The second phase training is noted by appending `*' to the architecture name. The `Rank' column represents the ranking of each method in the online score board at the time of submission.} 
\resizebox{\textwidth}{!}{%
\begin{tabular}{*{11}{c}}
\hline
Rank & Method  \multirow{2}*{ }&\multicolumn{3}{c}{Dice}&\multicolumn{3}{c}{Specificity }&\multicolumn{3}{c}{Sensitivity}\\
\cline{3-11}
 &  &Complete &Core &Enhancing &Complete &Core &Enhancing &Complete &Core &Enhancing\\
\hline
4&  \textsc{TwoPathCNN*}   &  0.85  &0.78  & 0.73  &0.93  &0.80  &0.72  &0.80  &0.76  &0.75   \\ 
9&   \textsc{LocalPathCNN*}  &  0.85  &0.74  &0.71  &0.91  &0.75  &0.71  &0.80  &0.77  &0.73   \\ 
10 &\textsc{AverageCNN*} &0.84  &0.75  &0.70  &0.95  &0.83  &0.73  &0.77  &0.74  &0.73 \\
14 &\textsc{GlobalPathCNN*} &0.82  &0.73  &0.68  &0.93  &0.81  &0.70  &0.75  &0.65  &0.70 \\
14&  \textsc{TwoPathCNN}  &  0.78  &0.63  &0.68  &0.67  &0.50  &0.59  &0.96  &0.89  &0.82  \\ 
15&  \textsc{LocalPathCNN}  &  0.77  &0.64  &0.68  &0.65  &0.52  &0.60  &0.96  &0.87  &0.80  \\ \hline
\end{tabular}
}

\label{tab:basicmodel}
\end{center}
\end{table*}

\subsection{Cascaded architectures}

We now discuss our experiments with the three cascaded architectures namely \textsc{InputCascadeCNN}, \textsc{LocalCascadeCNN} and \textsc{MFCascadeCNN}.  Table~\ref{tab:test} provides the quantitative results for each architecture. Figure~\ref{fig:results} also provides visual examples of the segmentation generated by each architecture.
\begin{table*}[t!]

\caption{Performance of the cascaded architectures. The reported results are from the second phase training. The `Rank' column shows the ranking of each method in the online score board at the time of submission.}

\begin{center}
\resizebox{\textwidth}{!}{%

\begin{tabular}{*{11}{c}}
\hline
Rank & Method  \multirow{2}*{ }&\multicolumn{3}{c}{Dice}&\multicolumn{3}{c}{Specificity}&\multicolumn{3}{c}{Sensitivity}\\

\cline{3-11}
 &  &Complete &Core &Enhancing &Complete &Core &Enhancing &Complete &Core &Enhancing\\
\hline

2&   \textsc{InputCascadeCNN*}&  0.88  &0.79  &0.73  &0.89  &0.79  &0.68  &0.87  &0.79  &0.80      \\ 
4-a&  \textsc{MFCascadeCNN*} &  0.86  &0.77  &0.73  &0.92  &0.80  &0.71  &0.81  &0.76  &0.76      \\ 
4-c&   \textsc{LocalCascadeCNN*} &  0.88  &0.76  &0.72  &0.91  &0.76  &0.70  &0.84  &0.80  &0.75      \\ \hline

\end{tabular}
}

\end{center}

\vspace*{-0.25cm}
\label{tab:test}
\end{table*}

We find that the \textsc{MFCascadeCNN*} model yields smoother boundaries between classes. We hypothesize that, since the neurons in the softmax output layer are directly connected to the previous outputs within each receptive field, these parameters are more likely to learn that the center pixel label should have a similar label to its surroundings. 

As for the \textsc{LocalCascadeCNN*} architecture, while it resulted in fewer false positives in the complete tumor category, the performance in other categories (i.e.\ tumor core and enhanced tumor) did not improve.

 Figure \ref{fig:multiple_views} shows segmentation results from the same brains (as in Figure~\ref{fig:results}) in Sagittal and Coronal views. The \textsc{InputCascadeCNN*} model was used to produce these results. As seen from this figure, although the segmentation is performed on Axial view but the output is consistent in  Coronal and Sagittal views. Although subjects in Figure 5 and Figure 6 are from our validation set for which the model is not trained on and the segmentation results from these subjects can give a good estimate of the models performance on a test set, however, for further clarity we visualise the models performance on two subjects from BRATS-2013 testst. These results are shown in Figure~\ref{fig:testresults} in Saggital (top) and Axial (bottom) views.
 
To better understand the process for which \textsc{InputCascadeCNN*} learns features, we present in Figure~\ref{fig:scheduled_prediction} the progression of the model by making predictions at every few epochs on a subject from our validation set. 

Overall, the best performance is reached by the \textsc{InputCascadeCNN*} model. It improves the Dice measure on all tumor regions. With this architecture, we were able to reach the second rank on the BRATS 2013 scoreboard. While \textsc{MFCascadeCNN*}, \textsc{TwoPathCNN*} and \textsc{LocalCascadeCNN*} are all ranked $4$, the inner ranking between these three models is noted as 4a, 4b and 4c respectively.  


Table~\ref{tab:brats-test} shows how our implemented architectures compare with currently published state-of-the-art methods as mentioned in \citep{Menze2014}\footnote{Please note that the results mentioned in Table~\ref{tab:brats-test} and Table~\ref{tab:leaderboard} are from methods competing in the BRATS 2013 challenge for which a static table is provided [\text{https://www.virtualskeleton.ch/BRATS/StaticResults2013}]. Since then, other methods have been added to the score board but for which no reference is available.}.  
The table shows that \textsc{InputCascadeCNN*} out performs Tustison et al.\, the winner of the BRATS 2013 challenge and is ranked first in the table.  Results from the BRATS-2013 leaderboard presented in Table~\ref{tab:leaderboard}  shows that our method outperforms other approaches on this dataset. We also compare our top performing method in Table~\ref{tab:brats12} with state-of-the-art methods on BRATS-2012, "4 label" test set as mentioned in \citep{Menze2014}. As seen from this table, our method out performs other methods in the tumor Core category and gets competitive results on other categories.

Let us mention that Tustison's method takes 100 minutes  to compute predictions per brain as reported in \citep{Menze2014}, while the \textsc{InputCascadeCNN*} takes 3 minutes, thanks to the fully convolutional architecture and the GPU implementation, which is over 30 times faster than the winner of the challenge. The \textsc{TwoPathCNN*}  has a performance close to the state-of-the-art. However, with a prediction time of 25 seconds, it is over 200 times faster than Tustison's method. Other top methods in the table are that of Meier et al and Reza et al with processing times of 6 and 90 minutes respectively. Recently \citet{Subbanna2014} published competitive results on the BRATS 2013 dataset, reporting dice measures of $0.86, 0.86, 0.77$ for Complete, Core and Enhancing tumor regions. Since they do not report Specificity and Sensitivity measures, a completely fair comparison with that method is not possible. However, as mentioned in~\citep{Subbanna2014}, their method takes 70 minutes to process a subject, which is about 23 times slower than our method. 

Regarding other methods using CNNs, \citet{Urban2014} used an average of two 3D convolutional networks with dice measures of $0.87, 0.77, 0.73$ for Complete, Core and Enhancing tumor regions on BRATS 2013 test dataset with a prediction time of about $1$ minute per model which makes for a total of $2$ minutes. Again, since they do not report Specificity and Sensitivity measures, we can not make a full comparison. However, based on their dice scores our \textsc{TwoPathCNN*} is similar in performance while taking only 25 seconds, which is four times faster. And the \textsc{InputCascadeCNN*} is better or equal in accuracy while having the same processing time. As for \citep{Zikic2014}, they do not report results on BRATS 2013 test dataset. However, their method is very similar to the \textsc{LocalPathCNN} which, according to our experiments, has worse performance.  

Using our best performing method, we took part in the BRATS 2015 challenge. The BRATS 2015 training dataset comprises of 220 subjects with high grade and 54 subjects with low grade gliomas. There are 53 subjects with mixed high and low grade gliomas for testing. Every participating group had 48 hours from receiving the test subjects to process them and submit their segmentation results to the online evaluation system. BRATS'15 contains the training data of 2013. The ground truth for the rest of the training brains is generated by a voted average of segmented results of the top performing methods in BRATS'13 and BRATS'12. Some of these automatically generated ground truths have been refined manually by a user. 

Because distribution of the intensity values in this dataset is very variable from one subject to another, we used a 7 fold cross validation for training. At test time, a voted average of these models was made to make prediction for each subject in the test dataset. The results of the challenge are presented in Figure~\ref{fig:brats15_results}. The semi-automatic methods participating in the challenge have been highlighted in grey. Please note since these results are not yet publicly available, we refrain from disclosing the name of the participants. In this figure the semi-automatic methods are highlighted in gray. As seen from the figure, our method ranks either first or second on Complete tumor and tumor Core categories and gets competitive results on active tumor category.  Our method has also less outliers than most other approaches.

\begin{figure}[t]
\centering

\includegraphics[width=0.98\linewidth]{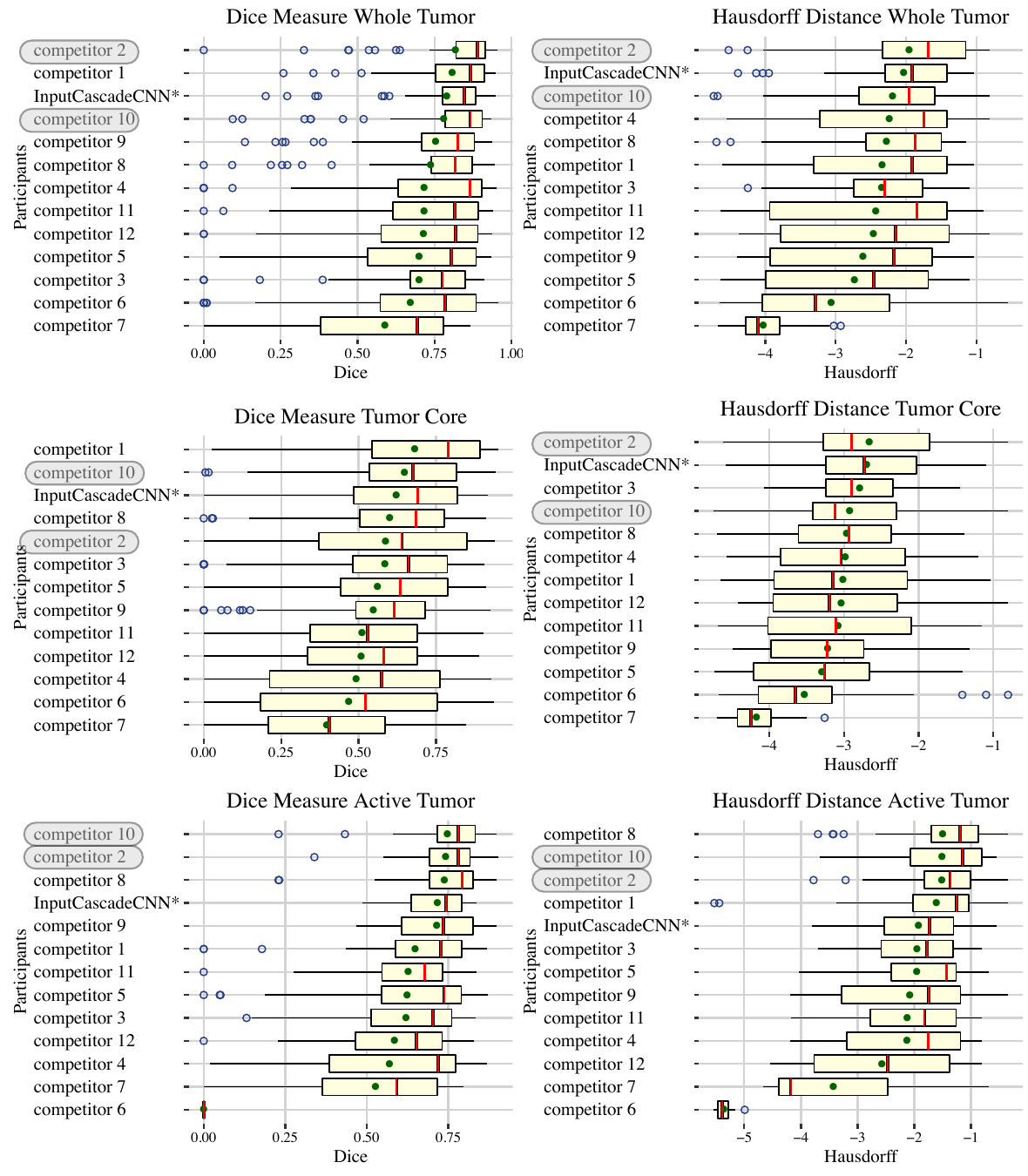}

\caption{Our BRATS'15 challenge results using \textsc{InputCascadeCNN*}. Dice scores and negative log Hausdorff distances are presented for the three tumor categories. Since the results of the challenge are not yet publicly available, we are unable to disclose the name of the participants. The semi-automatic methods are highlighted in gray. In each sub-figure, the methods are ranked based on the mean value. The mean is presented in green, the median in red and outliers in blue.}
\label{fig:brats15_results}
\end{figure}






\begin{table*}[tp]

\caption{Comparison of our implemented architectures with the state-of-the-art methods on the BRATS-2013 test set.
}

\begin{center}

\resizebox{\textwidth}{!}{%

\begin{tabular}{*{10}{c}}

\hline

Method \multirow{2}*{ }&\multicolumn{3}{c}{Dice}&\multicolumn{3}{c}{Specificity }&\multicolumn{3}{c}{Sensitivity}\\

\cline{2-10}

  &Complete &Core &Enhancing &Complete &Core &Enhancing &Complete &Core &Enhancing\\ \hline

 \rowcolor[gray]{0.9}\textsc{InputCascadeCNN*}&  0.88  &0.79  &0.73  &0.89  &0.79  &0.68  &0.87  &0.79  &0.80      \\ 

Tustison	&$0.87$	&$0.78$	&$0.74$	&$0.85$	&$0.74$	&$0.69$	&$0.89$	&$0.88$	&$0.83$	\\ 

 \rowcolor[gray]{0.9}\textsc{MFCascadeCNN*} &  0.86  &0.77  &0.73  &0.92  &0.80  &0.71  &0.81  &0.76  &0.76      \\ 


\rowcolor[gray]{0.9}\textsc{TwoPathCNN*} &0.85 & 0.78 & 0.73 &0.93  &0.80  &0.72  &0.80  &0.76  &0.75   \\ 

 \rowcolor[gray]{0.9}\textsc{LocalCascadeCNN*} &  0.88  &0.76  &0.72  &0.91  &0.76  &0.70  &0.84  &0.80  &0.75      \\ 

\rowcolor[gray]{0.9}\textsc{LocalPathCNN*}&  0.85  &0.74  &0.71  &0.91  &0.75  &0.71  &0.80  &0.77  &0.73   \\ 
Meier    	&$0.82$	&$0.73$	&$0.69$	&$0.76$	&$0.78$	&$0.71$	&$0.92$	&$0.72$	&$0.73$\\ 
Reza  	&$0.83$	&$0.72$	&$0.72$	&$0.82$	&$0.81$	&$0.70$	&$0.86$	&$0.69$	&$0.76$	\\ 
Zhao    	&$0.84$	&$0.70$	&$0.65$	&$0.80$	&$0.67$	&$0.65$	&$0.89$	&$0.79$	&$0.70$\\ 
Cordier 	&$0.84$	&$0.68$	&$0.65$	&$0.88$	&$0.63$	&$0.68$	&$0.81$	&$0.82$	&$0.66$\\ 
\rowcolor[gray]{0.9}\textsc{TwoPathCNN}&  0.78  &0.63  &0.68  &0.67  &0.50  &0.59  &0.96  &0.89  &0.82  \\ 
\rowcolor[gray]{0.9}\textsc{LocalPathCNN}&  0.77  &0.64  &0.68  &0.65  &0.52  &0.60  &0.96  &0.87  &0.80  \\ 
Festa 	&$0.72$	&$0.66$	&$0.67$	&$0.77$	&$0.77$	&$0.70$	&$0.72$	&$0.60$	&$0.70$	\\ 
Doyle 	&$0.71$	&$0.46$	&$0.52$	&$0.66$	&$0.38$	&$0.58$	&$0.87$	&$0.70$	&$0.55$	\\ \hline

\end{tabular}
}
\end{center}
\label{tab:brats-test}
\end{table*}

 \begin{table*}[]
\caption{ Comparison of our top implemented architectures with the state-of-the-art methods on the BRATS-2013 leaderboard set.
}
\begin{center}

\resizebox{\textwidth}{!}{%
\begin{tabular}{*{10}{c}}
\hline
Method \multirow{2}*{ }&\multicolumn{3}{c}{Dice}&\multicolumn{3}{c}{Specificity }&\multicolumn{3}{c}{Sensitivity}\\
\cline{2-10}
  &Complete &Core &Enhancing &Complete &Core &Enhancing &Complete &Core &Enhancing\\ \hline
\rowcolor[gray]{0.9}\textsc{InputCascadeCNN*}& 0.84& 	0.71& 0.57& 0.88& 	0.79& 	0.54& 	0.84& 	0.72& 	0.68 \\
Tustison	&0.79	&0.65	&0.53	&0.83	&0.70	&0.51	&0.81	&0.73	&0.66	\\
 Zhao    	&0.79	&0.59	&0.47	&0.77	&0.55	&0.50	&0.85	&0.77	&0.53   \\ 
Meier    	&0.72	&0.60	&0.53	&0.65	&0.62	&0.48	&0.88	&0.69	&0.6    \\ 
Reza  	    &0.73	&0.56	&0.51	&0.68	&0.64	&0.48	&0.79	&0.57	&0.63  	\\ 
Cordier 	&0.75	&0.61	&0.46	&0.79	&0.61	&0.43	&0.78	&0.72	&0.52   \\ \hline

\end{tabular}
}
\end{center}
\label{tab:leaderboard}
\end{table*}

 \begin{table}[]
\caption{ Comparison of our top implemented architectures with the state-of-the-art methods on the BRATS-2012 "4 label" test set as discussed in ~\cite{Menze2014}.
}
\begin{center}

\resizebox{0.5\textwidth}{!}{%
\begin{tabular}{*{4}{c}}
\hline
Method \multirow{2}*{ }&\multicolumn{3}{c}{Dice}\\
\cline{2-4}
  &Complete &Core &Enhancing\\ \hline
\rowcolor[gray]{0.9}\textsc{InputCascadeCNN*} &  	0.81& 0.72& 0.58	\\
Subbanna	&0.75	&0.70	&0.59	\\
 Zhao    	&0.82	&0.66	&0.42	\\ 
Tustison    	&0.75	&0.55	&0.52	\\ 
 Festa 	    &0.62	&0.50	&0.61	\\ 
 \hline

\end{tabular}
}
\end{center}
\label{tab:brats12}
\end{table}

\section{Conclusion}

In this paper, we presented an automatic brain tumor segmentation method based on deep convolutional neural networks.  We considered different architectures and investigated their impact on the performance. Results from the BRATS 2013 online evaluation system confirms that with our best model we managed to improve on the currently published state-of-the-art method both on accuracy and speed as presented in MICCAI 2013.  The high performance is achieved with the help of a novel two-pathway architecture (which can model both the local details and global context) as well as modeling local label dependencies by stacking two CNN's. 
Training is based on a two phase procedure, which we've found allows us to train CNNs efficiently when the distribution of labels is unbalanced.

Thanks to the convolutional nature of the models and by using an efficient GPU implementation, the resulting segmentation system is very fast. The time needed to segment an entire brain with any of the these CNN architectures varies between $25$ seconds and $3$ minutes, making them practical segmentation methods.



\section*{References}


{\small

\bibliographystyle{model2-names}

\bibliography{brats_cnn}

}

\end{document}